\documentclass[journal]{IEEEtran}
\usepackage[T1]{fontenc}
\usepackage{graphicx}
\usepackage{times}
\usepackage{helvet}
\usepackage{courier}
\usepackage{amsmath}
\usepackage{algorithm}
\usepackage{algorithmic}
\usepackage{csquotes} 
\usepackage{color}
\usepackage{paralist}
\usepackage{amssymb}
\usepackage{indentfirst}
\usepackage{subfigure}
\usepackage{float}
\usepackage{multirow}
\usepackage{cite}
\usepackage{mathrsfs}

\usepackage{mathrsfs} 
\usepackage{amsfonts}
\usepackage{todonotes}
\usepackage{pgfplots} 
\usepackage{epstopdf}
\usepackage{epsfig}
\pgfplotsset{compat=newest}
\usepackage{color}
\definecolor{forestgreen}{RGB}{0,139,69}

\usepackage{xcolor}
\definecolor{citecolor}{HTML}{0071bc}
\usepackage[colorlinks, linkcolor=red,  anchorcolor=blue, citecolor=citecolor]{hyperref} 

\usepackage{xcolor}
\definecolor{SeaGreen4}{RGB}{0,205,102} 
\definecolor{SlateBlue}{RGB}{106,90,205} 
\definecolor{DarkRed}{RGB}{178,34,34} 
	
\usepackage[switch]{lineno}

\usepackage{textcomp,booktabs}
\usepackage{amssymb}
\usepackage{pifont}
\newcommand{\cmark}{\ding{51}}%
\newcommand{\xmark}{\ding{55}}%

\usepackage{makecell}

\usepackage{colortbl}
\definecolor{mygray}{gray}{.9}
\definecolor{mypink}{rgb}{.99,.91,.95}
\definecolor{mycyan}{cmyk}{.3,0,0,0}

\begin{document}

\title{ Decoupling Amplitude and Phase Attention in Frequency Domain for RGB-Event based Visual Object Tracking } 

\author{Shiao Wang, Xiao Wang*, \emph{Member, IEEE}, Haonan Zhao, Jiarui Xu, 
    Bo Jiang*, \\ Lin Zhu, Xin Zhao, Yonghong Tian, \emph{Fellow, IEEE}, Jin Tang

\thanks{$\bullet$ Shiao Wang, Xiao Wang, Jiarui Xu, Bo Jiang, Jin Tang are with the School of Computer Science and Technology, Anhui University, Hefei 230601, China. (email: e24101001@stu.ahu.edu.cn, \{xiaowang, jiangbo, tangjin\}@ahu.edu.cn, 17809185626@163.com)} 

\thanks{$\bullet$ Haonan Zhao is with Northeastern University, Shenyang, China. (email: zhaohn@mails.neu.edu.cn)} 

\thanks{$\bullet$ Lin Zhu is with Beijing Institute of Technology, Beijing, China. (email: linzhu@pku.edu.cn)} 

\thanks{$\bullet$ Xin Zhao is with the School of Computer and Communication Engineering, University of Science and Technology Beijing. (email: xinzhao@ustb.edu.cn)} 

\thanks{$\bullet$ Yonghong Tian is with Peng Cheng Laboratory, Shenzhen, China; National Key Laboratory for Multimedia Information Processing, School of Computer Science, Peking University, China; School of Electronic and Computer Engineering, Shenzhen Graduate School, Peking University, China. (email: yhtian@pku.edu.cn) }

\thanks{* Corresponding Author: Xiao Wang, Bo Jiang} 
}

\markboth{ IEEE Transactions on ***, 2026 } 
{Shell \MakeLowercase{\textit{et al.}}: Bare Demo of IEEEtran.cls for IEEE Journals}

\maketitle

\begin{abstract}
Existing RGB–Event visual object tracking approaches primarily rely on conventional feature-level fusion, failing to fully exploit the unique advantages of event cameras. In particular, the high dynamic range and motion-sensitive nature of event cameras are often overlooked, while low-information regions are processed uniformly, leading to unnecessary computational overhead for the backbone network. 
To address these issues, we propose a novel tracking framework that performs early fusion in the frequency domain, enabling effective aggregation of high-frequency information from the event modality. Specifically, RGB and event modalities are transformed from the spatial domain to the frequency domain via the Fast Fourier Transform, with their amplitude and phase components decoupled. High-frequency event information is selectively fused into RGB modality through amplitude and phase attention, enhancing feature representation while substantially reducing backbone computation. In addition, a motion-guided spatial sparsification module leverages the motion-sensitive nature of event cameras to capture the relationship between target motion cues and spatial probability distribution, filtering out low-information regions and enhancing target-relevant features. Finally, a sparse set of target-relevant features is fed into the backbone network for learning, and the tracking head predicts the final target position. 
Extensive experiments on three widely used RGB–Event tracking benchmark datasets, including FE108, FELT, and COESOT, demonstrate the high performance and efficiency of our method. 
The source code of this paper will be released on \url{https://github.com/Event-AHU/OpenEvTracking}.  
\end{abstract}

\begin{IEEEkeywords}
Event Camera; RGB-Event Tracking; Frequency Fusion; Spatial Sparsification; Vision Transformer
\end{IEEEkeywords}

\IEEEpeerreviewmaketitle

\section{Introduction}

\IEEEPARstart{V}{isual} Object Tracking (VOT)~\cite{zhao2024biodrone, yao2025unctrack, zhang2025spiking, hu2022global, hu2024sotverse, chen2024crossei, zong2025enhancing, zhang2024multi} has long been a prominent research direction in the field of computer vision. In practical applications, conventional RGB cameras continue to be the dominant sensing modality, widely employed across diverse scenarios such as unmanned aerial vehicles, autonomous driving, intelligent surveillance, and other real-world settings. However, due to the inherent limitations of RGB cameras, such as their relatively low frame rate (typically 30 frames per second) and high sensitivity to illumination variations, they often exhibit unsatisfactory performance under extreme conditions, i.e., overexposure, low-light, and fast motion. These challenges can lead to motion blur and loss of critical information, significantly undermining the reliability of RGB-based tracking systems in dynamic environments. Consequently, researchers turn to alternative sensing modalities that complement and overcome the inherent limitations of traditional cameras, enabling more robust and effective visual tracking across a broader range of scenarios.

\begin{figure*}
\center
\includegraphics[width=\textwidth]{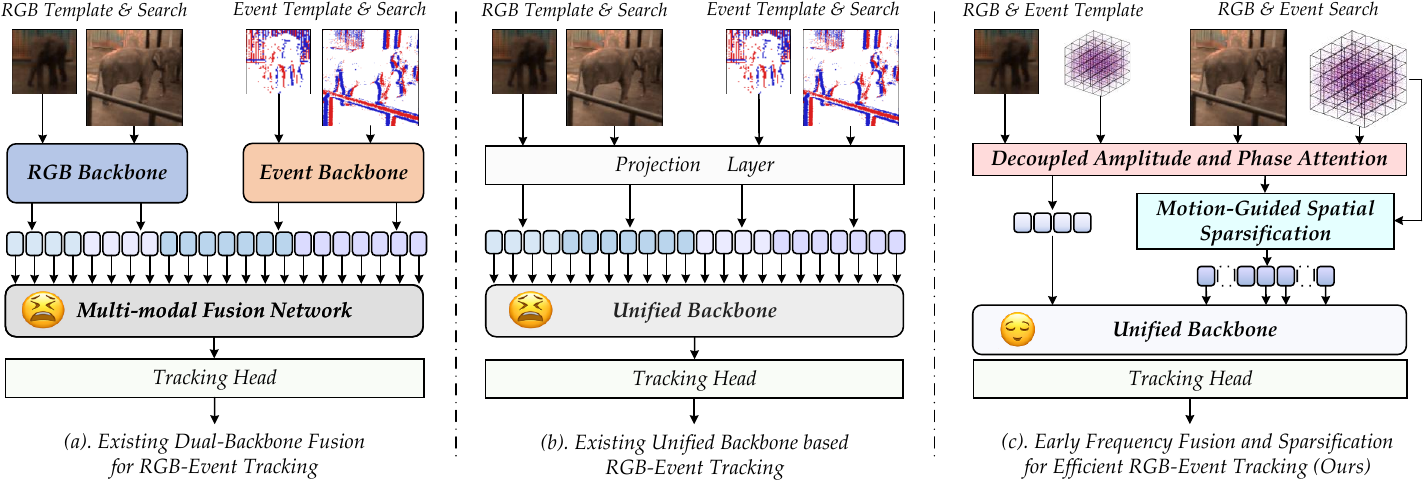}
\caption{(a, b) Traditional Siamese and single-stream trackers need to process all visual tokens during the multimodal feature fusion and extraction stages, respectively, leading to high computational complexity. (c) Our framework proposes \textbf{decoupled amplitude and phase attention} to halve the token count in early fusion, and uses \textbf{motion-guided spatial sparsification} to focus on target-relevant tokens, substantially reducing backbone computation.}  
\label{firstimg} 
\end{figure*}

Bio-inspired event cameras have attracted increasing attention from researchers~\cite{gehrig2024low, zubic2024state, yang2023event, chakravarthi2024recent} due to their high dynamic range and superior temporal resolution, enabling the capture of fast-moving scenes and subtle illumination changes that conventional frame-based cameras often fail to detect. By mimicking the human retina’s light perception mechanism, event cameras are highly sensitive to variations in scene brightness. An event signal with polarity (e.g., +1 or -1) is triggered only when the brightness of a scene increases or decreases beyond a predefined threshold. Unlike conventional RGB cameras, which output full image frames synchronously at a fixed frame rate, event cameras asynchronously record only brightness changes and generate corresponding event signals. As a result, event cameras impose minimal requirements on ambient illumination while capturing brightness variations with high temporal precision. Therefore, compared with traditional RGB cameras, they are particularly well-suited for tracking object motion in low-light or high-speed scenarios.

In recent years, an increasing number of tracking algorithms~\cite{tang2025revisiting, zhang2021object, zhu2023visual, zhu2023cross, zhang2023frame, wang2023visevent, huang2024mamba} have emerged, combining event cameras with RGB cameras to leverage the unique advantages of event cameras under extreme conditions and enhance the robustness of multimodal tracking. 
For example, Tang et al.~\cite{tang2025revisiting} preserve more temporal information by combining event voxels with RGB frames, and employ a vision Transformer~\cite{dosovitskiy2020image} for unified multimodal feature extraction and fusion. 
Zhang et al.~\cite{zhang2023frame} propose a high-frame-rate multimodal tracking framework that aligns and fuses RGB and event modalities, significantly improving the tracking performance.
However, existing RGB–Event multimodal tracking algorithms suffer from two major limitations:
\textit{(1) Challenges of Feature-level Fusion:} Most existing methods formulate RGB–Event tracking as a conventional multimodal fusion problem, focusing on achieving high tracking accuracy through feature-level fusion. Nevertheless, they often fail to effectively exploit the intrinsic characteristics of event data, namely its high dynamic range and temporal density.
\textit{(2) Limited Efficiency:} The joint processing of RGB frames and event streams substantially increases computational complexity.
As shown in Fig.~\ref{firstimg} (a) and (b), conventional Siamese trackers and widely used single-stream trackers often need to process all visual tokens at once when performing multimodal fusion or feature extraction, which significantly increases the computational burden of the network. Therefore, effectively integrating event modality features to achieve efficient RGB-Event visual object tracking remains challenging.

To address the aforementioned challenges, this work departs from conventional feature-level fusion by adopting early-stage modality fusion in the frequency domain, enabling selective aggregation of the complementary strengths of RGB and event modalities. As shown in Fig.~\ref{firstimg}(c), we introduce two core modules: the \textit{decoupled amplitude–phase attention module} and the \textit{motion-guided spatial sparsification module}. The first module leverages the high dynamic range of event cameras by employing an amplitude and phase attention aggregation method in the frequency domain. Specifically, the RGB and event modalities are first transformed from the spatial domain to the frequency domain, where their amplitude and phase components are decoupled. Using amplitude and phase attention, high-frequency information from the event modality is selectively integrated into the RGB modality, enhancing image quality under low illumination while simultaneously reducing the number of tokens input to the backbone by half.

For the motion-guided spatial sparsification module, a differential Transformer network based on the Fast Fourier Transform (FFT) extracts target-relevant motion information from the event voxels~\cite{zhu2018ev}. A lightweight score estimator then computes the spatial probability distribution of the target, while an exponential decay function determines an adaptive Top-$K$ value. This mechanism allows for the flexible selection of target-related tokens and suppression of background interference, further reducing the number of tokens processed by the backbone and enhancing target-focused feature representation.
Collectively, these two modules reduce computational cost and improve target-relevant feature extraction, enabling more effective RGB–Event tracking in challenging scenarios.

To sum up, the contributions of this work can be summarized as follows:

\textit{1).} We propose a novel amplitude and phase attention mechanism in the frequency domain, which aggregates high-frequency event information with RGB images at an early stage, thereby enhancing feature representations in challenging scenarios while significantly reducing the computational burden of the backbone.

\textit{2).} We introduce a motion-guided spatial sparsification strategy that selectively filters out redundant background information while enhancing target-relevant feature representations.

\textit{3).} Extensive experiments on three public datasets, i.e., FE108, FELT, and COESOT, fully demonstrate the effectiveness of the proposed multimodal tracker.

\section{Related Works} 


\subsection{RGB-Event based Tracking} 
Integrating RGB and event cameras to enhance object tracking performance has garnered significant interest within the research community.
In earlier work, 
Zhang et al.~\cite{zhang2021object} designed CDFI, which aligns frame and event representations and applies self- and cross-attention for robust tracking. 
Wang et al.~\cite{wang2023visevent} presented a baseline tracker using a cross-modality Transformer for effective feature fusion. 
Later works focus on the precise spatiotemporal alignment and cross-modal interaction. 
STNet~\cite{zhang2022spiking} is proposed to capture global spatial information and temporal cues by utilizing a Transformer and a spiking neural network (SNN). 
AFNet~\cite{zhang2023frame} adds an event-guided cross-modality alignment (ECA) module and a cross-correlation fusion head. 
Zhu et al.~\cite{zhu2023cross} further reduce modality conflict with orthogonal high-rank loss function and modality-masked tokens. 
Benchmark efforts unify RGB-Event training and evaluation for long sequences and diverse scenes.
Tang et al.~\cite{tang2025revisiting} provide a unified dataset and metric suite, and Wang et al.~\cite{wang2024long} release a long-term benchmark with a strong baseline. 
More recently, lightweight architectures and state-space-model (SSM) based trackers, such as Mamba-FETrack \cite{huang2024mamba}, have achieved a balance between model complexity and accuracy.
In parallel, ViPT~\cite{zhu2023visual}, SDSTrack \cite{hou2024sdstrack}, and EMTrack \cite{liu2024emtrack} allow for efficient parameter transfer by fine-tuning the trackers. 
Unlike the aforementioned works, we leverage the advantages of event cameras, using their high dynamic range to enhance the spatial structure representation of the RGB modality, while exploiting their high temporal resolution to integrate richer motion information.

\subsection{Frequency-Domain Modeling}
Frequency-domain modeling demonstrates a compelling paradigm for multimodal fusion. Specifically, classical correlation-filter trackers operate in the frequency domain using the Fast Fourier Transform (FFT). MOSSE~\cite{bolme2010visual} learns an adaptive filter with FFT-based optimization. KCF~\cite{henriques2015kcf} exploits a circulant structure and kernelization for fast dense sampling.

With the rise of deep learning, frequency-aware approaches have been widely explored to selectively amplify informative components while suppressing noise~\cite{jiang2021ffl, wang2021fcanet, liu2018mwcwnn, lee2021fnet,li2021fno}. For instance, Jiang et al.~\cite{jiang2021ffl} propose focal frequency loss that directs the model to focus on challenging spectral components. Wang et al.~\cite{wang2021fcanet} explore frequency channel attention by injecting FFT priors to refine channel-wise features. Meanwhile, FDA~\cite{yang2020fda} decouples spectra by swapping low-frequency amplitude across domains while keeping high-frequency phase. FFConv~\cite{Chi2020FFC} adds a spectral branch (FFT mixing iFFT) for long-range context. Subsequently, Kong et al.~\cite{Kong2023FreqDeblur} develop frequency-domain transformers with spectral attention and spectral feed-forward layers for deblurring, emphasizing sharp components and reducing computation. Chen et al.~\cite{Chen2024FADC} adapt the dilation rate to the local frequency and reweight frequency bands to preserve fine details in the segmentation task. FDConv~\cite{Chen2025FDConv} applies dynamic kernels per band to enhance structure and reduce noise for dense prediction. Zhang et al.~\cite{zhang2024dmfourllie} propose DMFourLLIE, a dual-stage multi-branch Fourier network that effectively mitigates color distortion and noise in low-light image enhancement. Cao et al.~\cite{cao2025exploiting} fuse event frames and grayscale frames in the Fourier domain to achieve effective action recognition. 
Compared with existing research, we fuse RGB and event in the frequency domain using amplitude and phase attention, and introduce a FFT-based differential ViT to enable target interaction between event templates and search regions.

\subsection{Event-based Motion Mining}  
Event cameras generate continuous event signals by asynchronously capturing changes in scene brightness, thereby achieving high temporal resolution and providing rich motion information. Capturing motion information from the event modality is essential.
Gallego et al.~\cite{gallego2018contrast} introduce a contrast maximization framework to directly recover camera or object motion from events. Their subsequent survey~\cite{gallego2020event} provides a comprehensive overview of motion-compensation approaches that establish stable priors for tracking.
End-to-end models like EV-FlowNet~\cite{zhu2018ev} learns optical flow from events in a self-supervised manner, turning sparse spikes into dense motion fields for low-latency alignment.  
E2VID ~\cite{rebecq2019e2vid} reconstructs high temporal resolution intensity frames from events, reducing motion blur and enabling reuse of frame-based modules. 
Liu et al.~\cite{liu2025edcflow} introduce an event-based optical flow estimation network, exploiting the complementarity of temporally dense motion features and cost-volume representations.
In the field of visual tracking, Zhang et al.~\cite{zhang2022spiking} proposed the Spiking Transformer, which treats the event stream as a continuous-time spike sequence with membrane dynamics.
CrossEI~\cite{chen2024crossei} proposes a motion-adaptive event sampling method and designs a bidirectional enhancement fusion framework to align and fuse event and image data.
In this work, we leverage event voxel representation to effectively preserve motion cues and further utilize these cues to guide the adaptive spatial sparsification of input tokens, enabling the model to focus on motion-relevant regions while suppressing redundant background information.

\begin{figure*}
\center
\includegraphics[width=\linewidth]{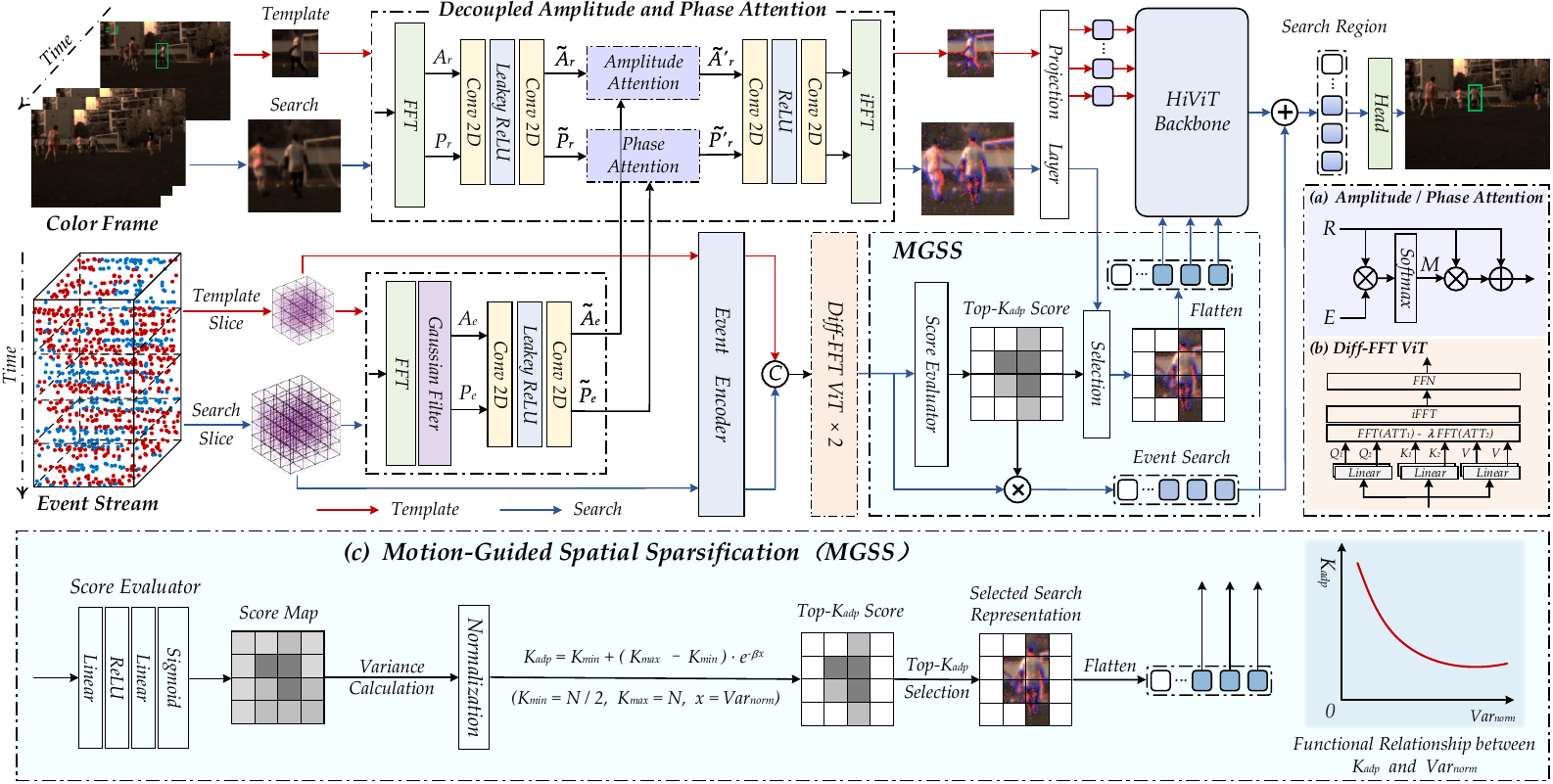}
\caption{An overview of our proposed \textbf{A}mplitude–\textbf{P}hase attention and \textbf{M}otion-guided sparsification framework for efficient RGB-Event tracking, called APMTrack. Specifically, RGB and event inputs are first decoupled into amplitude and phase in the frequency domain, allowing high-frequency event information to enhance RGB modality via amplitude and phase attention. The event encoder extracts motion cues, which are refined by the FFT-based differential ViT, and subsequently guide a spatial sparsification module for adaptive Top-$K$ token selection. The selected search tokens, combined with template features, are processed by the backbone, and the tracking head predicts the final target location.}
\label{framework}
\end{figure*}

\section{Our Proposed Approach} 

\subsection{Overview}
An overview of the proposed RGB-Event tracking framework is shown in Fig.~\ref{framework}, comprising two core modules. The \textit{Decoupled Amplitude and Phase Attention} module integrates high-frequency information from the event modality, which represents target contours, into the RGB modality in the frequency domain. This enhances the spatial structure representation of the RGB feature while reducing the computational load on the backbone network. The \textit{Motion-Guided Spatial Sparsification} module leverages motion information from the event modality. By modeling the temporal dynamics of sliced event voxels, it adaptively suppresses redundant background information while enhancing target-relevant feature representations. Together, these two modules fully exploit the high dynamic range and motion sensitivity of event cameras, enabling more efficient RGB-Event visual tracking. In the following sections, Section~\ref{input_representation} introduces the input representations, including RGB frames and event voxels. Section~\ref{network_architecture} details the two core modules, and Section~\ref{loss} presents the tracking head along with the loss function.

\subsection{Input Representation}
\label{input_representation}
Given an RGB video sequence with $N$ frames, denoted as \(I = \{I_1, I_2, \dots, I_N\}\) where $I_i \in \mathbb{R}^{3 \times H \times W}$ and $H$ and $W$ represent the spatial resolution of the camera, the corresponding asynchronous event stream can be represented as \(E = \{e_1, e_2, \dots, e_M\}\). Each event point $e_i$ is defined as a quadruple $\{x_i, y_i, t_i, p_i\}$, where $(x_i, y_i)$ denotes its spatial coordinates, $t_i \in [0, T]$ and $p_i \in \{-1,1\}$ denote the timestamp and polarity, respectively.

Following standard practice in visual tracking (e.g., OSTrack \cite{ye2022joint}), the template $Z_I \in \mathbb{R}^{3 \times H_z \times W_z}$ and search region $X_I \in \mathbb{R}^{3 \times H_x \times W_x}$ are obtained by cropping the first frame and subsequent frames at different scaling ratios from the RGB video sequence. 
For the event stream, to fully leverage its high temporal resolution and preserve rich motion-related information, we convert the event stream into an event voxel~\cite{zhu2018ev} sequence $V = \{V_1, V_2, \dots, V_N\}$, where $V_i \in \mathbb{R}^{B \times H \times W}$ represents $B$ time bins, constructed via spatiotemporal bilinear interpolation~\cite{Ye2023TowardsAO, ma2024timelens}. Similarly, the event template $Z_E \in \mathbb{R}^{B \times H_z \times W_z}$ and event search region $X_E \in \mathbb{R}^{B \times H_x \times W_x}$ are extracted from the event voxel sequence and aligned with the corresponding RGB frames.

\subsection{Network Architecture}
\label{network_architecture}

\noindent $\bullet$ \textbf{Preliminary: Fast Fourier Transform.~} 
First, we provide a brief introduction to the Fast Fourier Transform (FFT)~\cite{cooley1965algorithm}. FFT is an efficient algorithm for converting images from the spatial domain into the frequency domain. In this representation, the original image can be expressed as a superposition of sinusoidal waves at different frequencies, where each frequency component contains amplitude and phase information. The mathematical formulation of the FFT can be expressed as:

\begin{align}
\mathcal{F}(x)(u,v) &= X(u,v) \notag\\
&= \frac{1}{\sqrt{H \times W}} 
\sum_{h=0}^{H-1} \sum_{w=0}^{W-1} x(h,w) \, 
e^{-j 2 \pi \left( \frac{h u}{H} + \frac{w v}{W} \right)},
\label{2DFFT}
\end{align}
and its inverse transformation, i.e., the inverse Fast Fourier transform (iFFT), can be expressed as:
\begin{align}
\mathcal{F}^{-1}(X)(h,w) &= x(h,w) \notag\\
&= \frac{1}{\sqrt{H \times W}} 
\sum_{u=0}^{H-1} \sum_{v=0}^{W-1} 
X(u,v) \, e^{j 2 \pi \left( \frac{h u}{H} + \frac{w v}{W} \right)},
\label{iFFT}
\end{align}
where $x(h,w)$ and $X(u,v)$ denote the original image in the spatial domain and transformed representation in the frequency domain, respectively. Here, $h$ and $w$ represent the pixel coordinates along the height and width in the spatial domain, while $u$ and $v$ correspond to the vertical and horizontal frequency components in the frequency domain. $j$ is the imaginary unit. Each complex frequency component $X(u,v)$ can be decomposed into amplitude and phase, where the amplitude indicates the strength of the component in the image, while the phase determines its offset. These two components can be obtained by computing the magnitude and the angle of the complex component, respectively:
\begin{align}
\label{amplitude_phase}
\mathcal{A}(u,v) &= |X(u,v)| 
= \sqrt{R(X(u,v))^2 + I(X(u,v))^2}, \\
\mathcal{P}(u,v) &= \arg(X(u,v)) 
=  \arctan\frac{I(X(u,v))}{R(X(u,v))},
\end{align}
where $R(X(u,v))$ and $I(X(u,v))$ denote the real part and imaginary part of $X(u,v)$, respectively.

\noindent $\bullet$ \textbf{Decoupled Amplitude and Phase Attention.~} 
RGB frames and event data offer complementary representations, with RGB frames providing rich texture and color information, while event data captures motion cues and temporal changes. Direct spatial-domain fusion, however, may introduce blurring or misalignment owing to the inherent heterogeneity between the two modalities. By transforming both modalities into the frequency domain, we can selectively fuse RGB low-frequency structures with event high-frequency details, preserving both global appearance and fine temporal changes for a more robust representation. 
Therefore, we first transform the templates and search regions of both RGB and event data from the spatial domain to the frequency domain using FFT. Taking the search region $X_I$ and $X_E$ from the RGB frame and event voxel as examples, which can be expressed as:
\begin{align}
X_I^f = \mathbf{FFT}(X_I), \quad X_E^f = \mathbf{FFT}(X_E).
\end{align}
Subsequently, the complex frequency components are decomposed into amplitude and phase, where $A_r$ and $P_r$ represent the amplitude and phase of the RGB search. For event search, high-frequency information is first extracted via a Gaussian filter and then decoupled into $A_e$ and $P_e$. The amplitude and phase components of both modalities are then enhanced using 2D convolutions with the Leaky ReLU activation function, yielding $\tilde{A}_r$, $\tilde{P}_r$ for RGB, and $\tilde{A}_e$, $\tilde{P}_e$ for event modality. 

To effectively fuse RGB and event modalities in the frequency domain and incorporate the high-frequency information of events into RGB, we introduce amplitude and phase attention mechanisms for the corresponding branches. In each branch, both RGB and event amplitude/phase are L2-normalized along the channel dimension to stabilize the subsequent computations. As shown in Fig.~\ref{framework} (a), the features are then combined via element-wise multiplication, capturing interactions between the two modalities. A softmax function is applied to generate attention weights $M$, the RGB features are modulated by these weights $M$, with a residual connection applied, producing enhanced amplitude/phase features of RGB modality that incorporate high-frequency informative cues from the event modality. The process can be formulated mathematically as (the normalization operation is omitted):

\begin{align}
\tilde{A}_r' &= \mathbf{softmax} \Big( \tilde{A}_r \odot \tilde{A}_e \Big) \odot \tilde{A}_r + \tilde{A}_r, \\
\tilde{P}_r' &= \mathbf{softmax} \Big( \tilde{P}_r \odot \tilde{P}_e \Big) \odot \tilde{P}_r + \tilde{P}_r,
\end{align}
where $\tilde{A}_r'$ and $\tilde{P}_r'$ denote the fused amplitude and phase for RGB modality, with $\odot$ representing the element-wise multiplication operation.
After that, the fused amplitude and phase are converted into complex features and enhanced via Fast Fourier Convolution (FFC), implemented as a Convolution–ReLU–Convolution block. Finally, the enhanced features are restored from the frequency domain back to the spatial domain via the inverse Fast Fourier Transform (iFFT), yielding $X_I' \in \mathbb{R}^{B \times H_x \times W_x}$. Similarly, the RGB and event templates, i.e., $Z_I$ and $Z_E$, undergo the same process, producing $Z_I' \in \mathbb{R}^{B \times H_z \times W_z}$. Through the projection layer, $Z_I'$ and $X_I'$ are partitioned into non-overlapping patches and subsequently flattened into sequences $F_I^z \in \mathbb{R}^{N_z \times C}$ and $F_I^x \in \mathbb{R}^{N_x \times C}$, where $N_z$ and $N_x$ denote the number of tokens for the template and search region, respectively, and $C$ is the feature dimension.

The decoupled amplitude and phase attention module transforms the RGB and event modalities from the spatial domain to the frequency domain, decoupling them into amplitude and phase components. Through amplitude and phase attention, high-frequency motion information from the event modality is selectively integrated into the RGB modality. Only the fused RGB modality is retained as input to the backbone network, significantly reducing the computational cost.

\noindent $\bullet$ \textbf{Motion Representation Learning.~}
The motion information contained in asynchronous event streams reflects the target's state in the current scene. Accordingly, we model the event motion to capture the spatial distribution patterns triggered by the target's movement across distinct scenarios. We represent the event stream using event voxel accumulated within temporal windows, denoted as $Z_E \in \mathbb{R}^{B \times H_z \times W_z}$ for event template and $X_E \in \mathbb{R}^{B \times H_x \times W_x}$ for event search. This representation encodes the event distribution over $B$ temporal bins, where each bin stores the spatial activation of events within its corresponding time interval. 

To extract the feature representations of these temporal bins, both the event template and event search are fed into an event encoder. This encoder is composed of multi-scale Conv-BN-LeakyReLU blocks followed by linear projection layers, enabling the extraction of spatiotemporal features $Z_E' \in \mathbb{R}^{B \times \frac{H_z}{16} \times \frac{W_z}{16} \times C}$ and $X_E' \in \mathbb{R}^{B \times \frac{H_x}{16} \times \frac{W_x}{16} \times C}$ for the event template and search (Note that, to align the event sequence length with that of the RGB sequence, here $\frac{H_z}{16} \times \frac{W_z}{16}$ = $N_z$ and $\frac{H_x}{16} \times \frac{W_x}{16}$ = $N_x$). To capture continuous motion features, we warp the event feature maps to align with the RGB reference and then compute the dense difference maps. Specifically, we first warp the features $Z_{E_i}' \in \mathbb{R}^{\frac{H_z}{16} \times \frac{W_z}{16} \times C}$ and $X_{E_i}' \in \mathbb{R}^{\frac{H_x}{16} \times \frac{W_x}{16} \times C}$, $i=1,2,\ldots, B$, of temporal bins towards RGB template and search feature maps through bilinear interpolation, and obtain $\tilde{Z}_{E_i}' \in \mathbb{R}^{\frac{H_z}{16} \times \frac{W_z}{16} \times C}$ and $\tilde{X}_{E_i}' \in \mathbb{R}^{\frac{H_x}{16} \times \frac{W_x}{16} \times C}$, $i=1,2,\ldots, B$. Subsequently, we introduce a sampling stride $s=1$ to control the temporal interval between consecutive bins, yielding feature difference maps that reflect dense spatial displacements. The formulation is as follows: 

\begin{align}
    D_j^z &= \tilde{Z}_{E_{j+1}}' - \tilde{Z}_{E_{j}}', \quad j = 1, \ldots, B - 1, \\
    D_j^x &= \tilde{X}_{E_{j+1}}' - \tilde{X}_{E_{j}}', \quad j = 1, \ldots, B - 1,
\end{align}
where $D_j^z \in \mathbb{R}^{\frac{H_z}{16} \times \frac{W_z}{16} \times C}$ and $D_j^x \in \mathbb{R}^{\frac{H_x}{16} \times \frac{W_x}{16} \times C}$ represent the $j$-th difference maps of the event template and search region, respectively, capturing the motion evolution of events across consecutive temporal bins $(j+1)$-th and $j$-th.

\noindent $\bullet$ \textbf{Motion-Guided Spatial Sparsification.~}
In the following, we detail how motion difference maps can be leveraged to perform spatial sparsification under diverse scene conditions. To exploit the dynamic frequency information in event signals, we introduce an FFT-based differential Transformer module (Diff-FFT ViT), which enables effective interaction between the event template and the search region to capture dynamic cues associated with the target object, as shown in Fig.~\ref{framework}(b). Inspired by~\cite{ye2024differential}, which reduces noise by computing the difference between two independent softmax attention maps, we adopt a similar differential strategy in our design. Specifically, the multi-time-scale difference maps of the template and the search region are first concatenated, followed by global average pooling, yielding the global motion representations $D_z \in \mathbb{R}^{N_z \times C}$ and $D_x\in \mathbb{R}^{N_x \times C}$, respectively. Then, we concatenate the global motion features of the template and the search region to obtain $D \in \mathbb{R}^{N \times C}$ (here $N = N_z+N_x$), which is subsequently fed into the Diff-FFT ViT module. Given an input $D \in \mathbb{R}^{N \times C}$, we first project it to the queries, keys, and values:
$Q_1, Q_2, K_1, K_2 \in \mathbb{R}^{N \times \frac{C}{2}}$ and $V \in \mathbb{R}^{N \times C}$:

\begin{align}
    [Q_1; Q_2] = D W_Q, \quad
    [K_1; K_2] = D W_K, \quad
    V = D W_V,
\end{align}
where $W_Q,W_K,W_V$ are learnable projection matrices. Then, the FFT-based differential attention operator computes outputs via:

\begin{align}
    ATT_1 &= \mathbf{softmax}\Big(\frac{Q_1 K_1^\top}{\sqrt{d}}\Big) V, \\
    ATT_2 &= \mathbf{softmax}\Big(\frac{Q_2 K_2^\top}{\sqrt{d}}\Big) V, 
\end{align}

\begin{align}
    D_{fft} = \Big(\, \mathbf{FFT}(ATT_1) - \lambda \, \mathbf{FFT}(ATT_2)\,\Big),
\end{align}
where $D_{fft}$ is the output of the Diff-FFT attention operator in the frequency domain, and $\lambda$ is a learnable scalar. Afterwards, we apply the inverse FFT (iFFT) to transform the output from the frequency domain back to the spatial domain as follows:

\begin{align}
    D_s = \mathbf{iFFT}(D_{fft} \odot f),
\end{align}
where $f$ denotes the Gaussian window, which attenuates insignificant frequency components to further reduce the impact of noise on the resulting attention. Finally, the feed-forward network (FFN) is applied to further enhance the feature representations, mixing information across channels and outputting the target-enhanced motion representation $D' \in \mathbb{R}^{N \times C}$. 

After obtaining the target-enhanced motion features, we extract the search region component $D'_x \in \mathbb{R}^{N_x \times C}$ and feed it into the MGSS (motion-guided spatial sparsification) module to guide spatial sparsification. As shown in Fig~\ref{framework} (c), a score estimator, consisting of a multi-layer perceptron (MLP), is initially employed to project the motion representation of the search region into a corresponding score map $\mathbf{S} \in \mathbb{R}^{N_x}$, where each element $S_{i} \in [0,1]$ represents the probability score of the corresponding position belonging to the target-relevant motion region. We then calculate and normalize the variance of the score map to characterize the spatial distribution pattern of target motion-guided information in the current scene. Higher variance indicates that target-relevant information is concentrated in a few salient regions, whereas lower variance indicates a more spatially scattered distribution. Therefore, when the variance is low, a greater number of spatial regions must be preserved to adequately represent target information, whereas high variance permits more redundant background regions to be safely filtered out. Guided by this principle, we propose a variance-based adaptive Top-$K$ function that dynamically determines an appropriate $K_{adp}$ based on the scene-wise variance of target probabilities, enabling the adaptive pruning of redundant low-scoring regions. The functional relationship between $K_{adp}$ and $Var_{norm}$ is defined as:

\begin{align}
    K_{adp} = K_{min} + (K_{max}-K_{min}) \cdot e^{- \beta x}, \quad (\beta \in \mathbb{N}^+)
\end{align}
where $K_{min}$ and $K_{max}$ are used to determine the lower and upper bounds of $K_{adp}$, and $x$ denotes the normalized variance $V_{norm}$. $\beta \in \mathbb{N}^+$ is a hyperparameter. This exponential mapping enables smooth and adaptive control over the retained token ratio, preserving more high-scoring regions in complex scenes while effectively pruning redundancy when information is concentrated. Additional comparisons with alternative formulations are provided in the ablation experiments.

Subsequently, we perform a Top-$K$ selection on $F_I^x$, retaining the $K$ highest-scoring patches to obtain the selected search feature representation $F_I^{\text{top}} \in \mathbb{R}^{N_x^{\text{top}} \times C}$, where $N_x^{\text{top}}$ denotes the number of retained tokens. This representation is then concatenated with the template feature $F_I^z$ and fed into the HiViT~\cite{zhang2023hivit} backbone network for further feature learning and interaction. Similarly, a Top-$K$ selection is applied to the event search region, and the selected event search features are multiplied element-wise with the corresponding Top-$K$ scores, enhancing the relative importance of motion representations across different regions. Finally, the Top-$K$ search region features from both modalities are combined via summation and then, after zero-padding the spatially redundant regions, fed into the tracking head to obtain the final tracking predictions.

The motion-guided spatial sparsification module exploits the intrinsic motion sensitivity of event cameras to capture dynamic states across diverse scenes. It employs an FFT-based differential ViT to model target-related motion cues. A variance-driven function of the target probability distribution is used to identify the Top-$K$ salient regions in each scene, enabling selective suppression of redundant spatial areas while preserving target-focused information. This design enhances target-related feature representation while reducing computational overhead.

\subsection{Tracking Head and Loss Function} 
\label{loss}

Our tracking head design follows OSTrack \cite{ye2022joint}. The enhanced search region features extracted from the ViT backbone are fed into the tracking head to predict the spatial location of the target. Specifically, the search features are first reshaped into a 2D feature map, which is subsequently processed by a series of Convolution–Batch Normalization–ReLU (Conv–BN–ReLU) blocks, and then produces four outputs: (1) a target classification score map indicating the probability of the target appearing at each spatial location; (2) local offset predictions for refining the estimated center coordinates of the bounding box; (3) normalized bounding box dimensions (width and height); and (4) the final predicted bounding box, representing the estimated target location in the current frame.

During training, we adopt a loss formulation similar to OSTrack~\cite{ye2022joint}, employing three distinct loss functions for comprehensive optimization: Focal Loss (\(L_{focal}\)) for target classification, L1 Loss (\(L_1\)) for offset regression, and GIoU Loss (\(L_{GIoU}\)) for bounding box size and overlap regression. The overall training objective \(L_{total}\) is defined as a weighted combination of these terms:
\begin{equation}
    L_{total} = \lambda_1 L_{focal} + \lambda_2 L_1 + \lambda_3 L_{GIoU},
    \label{eq:total_loss}
\end{equation}
where \(\lambda_1, \lambda_2, \lambda_3\) are weighting coefficients balancing the contribution of each loss term.

\section{Experiments} 

\subsection{Datasets and Evaluation Metric}

In this section, we compare with other state‑of‑the‑art (SOTA) trackers on existing event-based tracking datasets, including \textbf{FE108}~\cite{zhang2021object}, \textbf{FELT}~\cite{wang2024long}, and \textbf{COESOT}~\cite{tang2025revisiting}.

\noindent $\bullet$ \textbf{FE108 Dataset}: The dataset is a dual-modal single-object tracking benchmark collected using a grayscale DAVIS 346 event camera. It contains 108 video clips recorded in indoor environments, with a total duration of approximately 1.5 hours. Among them, 76 videos are used for training and 32 for testing. The dataset covers 21 categories of objects, which can be grouped into three types: animals, vehicles, and everyday items. In addition, the dataset includes four challenging scene conditions: low illumination (LL), high dynamic range (HDR), and fast-motion scenes where motion blur is either present or absent in APS frames (FWB and FNB). Please refer to the following GitHub for more details \url{https://github.com/Jee-King/ICCV2021_Event_Frame_Tracking?tab=readme-ov-file}.

\begin{table*}[h]
\centering
\small
\caption{Experimental results (SR/PR) on FE108 dataset.}
\label{tab:fe108_results}
\begin{tabular}{cccccccc}
\toprule
\textbf{SiamRPN~\cite{li2018high}} & \textbf{SiamBAN~\cite{chen2020siamese}} & \textbf{SiamFC++~\cite{xu2020siamfc++}} & \textbf{KYS~\cite{bhat2020know}} & \textbf{CLNet~\cite{dong2020clnet}} & \textbf{CMT-MDNet~\cite{wang2023visevent}} & \textbf{ATOM~\cite{danelljan2019atom}} \\
21.8/33.5 & 22.5/37.4 & 23.8/39.1 & 26.6/41.0 & 34.4/55.5 & 35.1/57.8 & 46.5/71.3 \\ 
\hline
\textbf{DiMP~\cite{bhat2019learning}} & \textbf{PrDiMP~\cite{danelljan2020probabilistic}} & \textbf{CEUTrack~\cite{tang2025revisiting}} & \textbf{FENet~\cite{zhang2021object}}  & \textbf{ViPT~\cite{zhu2023visual}}  & \textbf{MamTrack~\cite{sun2025exploring}} & \textbf{Ours} \\
52.6/79.1 & 53.0/80.5 & 55.6/84.5 & 63.4/92.4   & 65.2/92.1     &\textbf{66.4}/94.2 & 64.4/\textbf{95.2} \\
\bottomrule
\end{tabular}
\end{table*}

\noindent $\bullet$ \textbf{FELT Dataset}: The dataset is a large-scale, long-duration dual-modal single-object tracking benchmark collected using a DAVIS 346 event camera. It comprises 1,044 video sequences, each with an average duration of over 1.5 minutes, ensuring that every video contains at least 1,000 pairs of synchronized RGB and event frames. Among them, 730 sequences are used for training and 314 for testing. In total, the dataset provides 1,949,680 annotated frames, covering 60 object categories and defining 14 challenging tracking attributes, including occlusion, fast motion, and low-light conditions. Please refer to the following GitHub for more details \url{https://github.com/Event-AHU/FELT_SOT_Benchmark}. 

\noindent $\bullet$ \textbf{COESOT Dataset}: 
This benchmark dataset is an RGB-Event-based, category-wide tracking dataset designed to evaluate the generalization ability of tracking algorithms across different object types. It contains 1,354 video sequences covering 90 object categories, with 827 sequences used for training and 527 for testing, providing a total of 478,721 annotated RGB frames. To facilitate fine-grained analysis of algorithm performance, the dataset explicitly defines 17 challenging factors, including fast motion, occlusion, illumination variation, background clutter, and scale changes. Please refer to the following GitHub for more details \url{https://github.com/Event-AHU/COESOT}.

\begin{table}
\centering
\small
\setlength{\tabcolsep}{4pt} 
\caption{Experimental results on FELT dataset.} 
\label{tab:felt_results}
\begin{tabular}{l|c|ccc|c}
\toprule
\textbf{Trackers} & \textbf{Source} &\textbf{SR}  &\textbf{PR} &\textbf{NPR} &\textbf{FPS} \\
\hline 
\textbf{01. STARK~\cite{yan2021learning}} & ICCV21 & 52.7 & 67.9 & 62.8 & 42 \\ 
\textbf{02. OSTrack~\cite{ye2022joint}} & ECCV22 & 52.3 & 65.9 & 63.3 & 75 \\ 
\textbf{03. MixFormer~\cite{Cui2022MixFormerET}} & CVPR22 & 53.0 & 67.5 & 63.8 & 36 \\ 
\textbf{04. AiATrack~\cite{gao2022aiatrack}} & ECCV22 & 52.2 & 66.7 & 62.8 & 31 \\ 
\textbf{05. SimTrack~\cite{chen2022backbone}} & ECCV22 & 49.7 & 63.6 & 59.8 & 82 \\ 
\textbf{06. GRM~\cite{Gao2023GeneralizedRM}} & CVPR23  & 52.1 & 65.6 & 62.9 & 38 \\ 
\textbf{07. ROMTrack~\cite{cai2023robust}} & ICCV23  & 51.8 & 65.8 & 62.7 & 64 \\ 
\textbf{08. ViPT~\cite{zhu2023visual}} & CVPR23 & 52.8 & 65.3 & 63.1 & 29  \\ 
\textbf{09. SeqTrack~\cite{chen2023seqtrack}} & CVPR23  & 52.7 & 66.9 & 63.4 & 31 \\ 
\textbf{10. ARTrackv2~\cite{bai2024artrackv2}} & CVPR24 & 52.3 & 65.2 & 62.8 & 42 \\ 
\textbf{11. HIPTrack~\cite{cai2024hiptrack}} & CVPR24 & 51.6 & 65.6 & 62.2 & 39 \\ 
\textbf{12. ODTrack~\cite{Zheng2024ODTrackOD}} & AAAI24 & 52.2 & 66.0 & 63.5 & 57 \\ 
\textbf{13. EVPTrack~\cite{shi2024explicit}} & AAAI24 & 53.8 & 68.7 & 64.8 & 45 \\ 
\textbf{14. AQATrack~\cite{xie2024autoregressive}} & CVPR24 & 54.0 &69.1 & 64.7 & 41 \\ 
\textbf{15. SDSTrack~\cite{hou2024sdstrack}} & CVPR24 & 53.7 & 66.4 & 64.1 & 28 \\ 
\textbf{16. UnTrack~\cite{wu2024single}} & CVPR24 & 53.6 & 66.0 & 63.9 & 12 \\ 
\textbf{17. FERMT~\cite{zheng2024exploring}} & ECCV24 & 51.8 & 66.1 & 62.9 & 70 \\ 
\textbf{18. LMTrack~\cite{xu2025less}} & AAAI25 & 50.9 & 63.9 & 61.8 & 72 \\ 
\textbf{19. AsymTrack~\cite{zhu2025two}} & AAAI25 & 51.9 & 66.7 & 62.0 & 104 \\ 
\textbf{20. SUTrack~\cite{chen2025sutrack}} & AAAI25 &\textbf{56.6} &70.9 &66.6  & 25 \\ 
\textbf{21. ORTrack~\cite{wu2025learning}} & CVPR25 & 48.4 & 61.7 & 59.2 & 90 \\ 
\textbf{22. UNTrack~\cite{qin2025must}} & CVPR25 & 50.0 & 63.9 & 61.6 & 23 \\ 
\hline
\textbf{23. Ours} & - &56.5 &\textbf{72.3}  &\textbf{67.9} &27  \\ 
\bottomrule
\end{tabular} 
\end{table}

\begin{table}
\centering
\small
\caption{Tracking results on COESOT Dataset.} 
\label{coesot_result}
\begin{tabular}{l|c|cc}
\hline \toprule [0.5 pt]
\textbf{Trackers} &\textbf{Source}  &\textbf{SR}  &\textbf{PR} \\
\hline 
\textbf{01. TransT~\cite{Chen2021TransformerT}} & CVPR21  &60.5   &72.4   \\
   \textbf{02. STARK~\cite{yan2021learning}} & ICCV21  &56.0 &67.7 \\
   \textbf{03. OSTrack~\cite{ye2022joint}} & ECCV22  &59.0  &70.7 \\
   \textbf{04. MixFormer~\cite{Cui2022MixFormerET}} & CVPR22  &55.7 &66.3 \\
   \textbf{05. AiATrack~\cite{gao2022aiatrack}} & ECCV22 &59.0 &72.4  \\
   \textbf{06. SiamR-CNN~\cite{voigtlaender2020siam}} & CVPR20  &60.9  &71.0     \\
   \textbf{07. ToMP50~\cite{Mayer2022TransformingMP}} & CVPR22  &59.8 &70.8 \\
   \textbf{08. ToMP101~\cite{Mayer2022TransformingMP}} & CVPR22 &59.9 &71.6 \\
   \textbf{09. KeepTrack~\cite{mayer2021learning}} & ICCV21  &59.6  &70.9  \\
   \textbf{10. PrDiMP50~\cite{danelljan2020probabilistic}} & CVPR20  &57.9 &69.6  \\
   \textbf{11. DiMP50~\cite{bhat2019learning}} &ICCV19  &58.9 &72.0 \\
   \textbf{12. ATOM~\cite{danelljan2019atom}} &CVPR19   &55.0 &68.8  \\
   \textbf{13. TrDiMP~\cite{wang2021transformer}} &CVPR21  &60.1 &72.2  \\
   \textbf{14. MDNet~\cite{wang2023visevent}} &TCYB23   &53.3 &66.5 \\
   \textbf{15. ViPT~\cite{zhu2023visual}} &CVPR23   &\textbf{68.3} &81.0 \\
   \textbf{16. SDSTrack~\cite{hou2024sdstrack}} & CVPR24 & 66.7 & 79.7 \\
   \textbf{17. UnTrack~\cite{wu2024single}} & CVPR24 & 67.9 & 80.9 \\
   \textbf{18. CEUTrack~\cite{tang2025revisiting}}   &PR25 & 62.7 & 76.0   \\
   \textbf{19. LMTrack~\cite{xu2025less}}   &AAAI25 & 58.4 & 71.1   \\
   \textbf{20. CMDTrack~\cite{zhang2025cross}} &TPAMI25   &65.7 &74.8 \\
   \textbf{21. MCITrack~\cite{kang2025exploring}}   &AAAI25 & 64.7 & 78.1   \\
\hline
   \textbf{22. Ours} &- &68.0 &\textbf{83.3} \\
\hline \toprule [0.5 pt]
\end{tabular} 
\end{table}

For evaluation, three commonly used metrics are employed: \textbf{Precision (PR)}, \textbf{Normalized Precision (NPR)}, and \textbf{Success Rate (SR)}. Specifically, Precision (PR) measures the proportion of frames where the distance between the predicted and ground-truth centers is below a predefined threshold (default: 20 pixels). Normalized Precision (NPR) computes the Euclidean distance between the predicted and ground-truth centers and normalizes it using the diagonal matrix formed by the width and height of the ground-truth bounding box. Success Rate (SR) represents the proportion of frames in which the Intersection over Union (IoU) between the predicted and ground-truth bounding boxes exceeds a specified threshold.

\subsection{Implementation Details} 

We adopt HiViT~\cite{zhang2023hivit} as the backbone network for deep feature interaction learning, initialized with the pretrained weights of SUTrack-B224~\cite{chen2025sutrack}. Each event voxel is discretized into 5 time bins within each time window. The learning rate is set to 0.0001, and the weight decay is 0.0001. The model is trained for 50 epochs, with 60,000 template–search pairs per epoch, and a batch size of 32. The input resolution of the template and search region is fixed to $112 \times 112$ and $224 \times 224$, respectively. For the adaptive Top-$K$ function, we set $K_{min}=\frac{N_x}{2}$ and $K_{max}=N_x$ in this work, while the hyperparameter $\beta$ is set to 2.

We use AdamW~\cite{loshchilov2018adamw} as the optimizer, where the loss weighting coefficients $\lambda_i$ ($i = 1,2,3$) are set to 1, 5, and 2, respectively. Our implementation is based on Python and PyTorch~\cite{paszke2019pytorch}. All experiments are conducted on a server equipped with an AMD EPYC 7542 32-core CPU and an NVIDIA RTX 4090 GPU. Further implementation details are provided in the released source code.

\begin{table*} 
\center
\caption{Component Analysis on the COESOT Dataset.} 
\label{CAResults} 
\begin{tabular}{c|ccccc|cc|cc} 		
\hline \toprule 
\textbf{No.}  &\textbf{Concatenation}  &\textbf{Amplitude-Attn} &\textbf{Phase-Attn}  &\textbf{Diff-FFT ViT}   &\textbf{MGSS} &\textbf{SR}  &\textbf{PR} &\textbf{Params} &\textbf{FLOPs} \\  
\hline 
$\#1$ &\cmark   &\xmark    &\xmark  &\xmark   &\xmark  &66.9  &81.9  &74.1M   &1167.4G  \\
\hline
$\#2$ &\xmark   &\cmark    &\xmark  &\xmark   &\xmark  &66.8  &81.7  &70.0M   &608.9G  \\
$\#3$ &\xmark   &\xmark    &\cmark  &\xmark   &\xmark  &66.5  &81.3  &70.0M   &608.9G  \\
$\#4$ &\xmark   &\cmark    &\cmark  &\xmark   &\xmark  &67.2  &82.4  &70.0M   &608.9G  \\
$\#5$ &\xmark   &\cmark    &\cmark  &\xmark   &\cmark  &67.6  &82.9  &73.3M   &726.0G  \\
$\#6$ &\xmark   &\cmark    &\cmark  &\cmark   &\cmark  & \textbf{68.0} & \textbf{83.3}   &77.0M   &701.0G  \\
\hline \toprule  
\end{tabular}
\end{table*}

\subsection{Comparison on Public Benchmark Datasets}

\noindent $\bullet$ \textbf{Results on FE108 Dataset.~}
As shown in Table~\ref{tab:fe108_results}, we conduct a performance comparison on the FE108 dataset between our proposed method and several other SOTA approaches. The experiment results indicate that our method achieves superior performance, with SR and PR scores of 64.4 and 95.2, respectively. Although the SR metric is slightly lower than the best, we achieve the highest PR score, which indicates that our model offers more accurate target center localization. We attribute this to the fact that our framework emphasizes modeling the key target regions, leading to more precise localization, while boundary and scale estimation still have room for further improvement.

\noindent $\bullet$ \textbf{Results on FELT Dataset.~}
As shown in Table~\ref{tab:felt_results}, we evaluate our method on the FELT dataset and compare it with comprehensive SOTA visual trackers. Our approach achieves an SR of 56.5, a PR of 72.3, and an NPR of 67.9, achieving a new SOTA performance. It outperforms recent strong baselines such as AsymTrack~\cite{zhu2025two} and AQATrack~\cite{xie2024autoregressive}, demonstrating superior tracking accuracy and stability. Moreover, compared with the second-best method, SUTrack~\cite{chen2025sutrack}, our approach attains improvements of +1.4 in PR and +1.3 in NPR, while achieving comparable performance in SR. These results confirm that our framework effectively exploits event-guided motion cues, delivering strong robustness under challenging long-term tracking conditions.

\noindent $\bullet$ \textbf{Results on COESOT Dataset.~}
We also report the comparison results on the COESOT dataset, as shown in Table~\ref{coesot_result}. Our method achieves an SR score of 68.0, ranking among the top methods and second only to ViPT~\cite{zhu2023visual} (68.3). More notably, our approach achieves a PR score of 83.3, significantly outperforming all other trackers, including recent SOTA methods such as UnTrack~\cite{wu2024single} (80.9) and SDSTrack~\cite{hou2024sdstrack} (79.7). This substantial improvement in precision clearly demonstrates the effectiveness and robustness of our tracker. Overall, the results validate that our approach achieves highly competitive performance across various categories, establishing a new SOTA in terms of tracking precision on the COESOT benchmark.


\subsection{Component Analysis}
As presented in Table~\ref{CAResults}, we analyze the core components of our framework individually, highlighting the contribution and necessity of each module. To begin with, we evaluate the feature-level concatenation strategy as a comparative baseline (first row), yielding SR and PR scores of 66.9 and 81.9, respectively. In rows 2-4, we decouple the amplitude and phase of the two modalities and evaluate three settings: amplitude-only attention, phase-only attention, and both jointly. The results show that using amplitude or phase alone leads to performance degradation, whereas combining both yields better results. In the fifth row, we introduce the MGSS (Motion-Guided Spatial Sparsification) module, which removes redundant regions and enhances the model’s focus on the target area, thereby improving accuracy. In the last row, we further incorporate the FFT-based Differential ViT, enabling more effective modeling of target-related motion cues. Finally, our framework achieves the best overall performance, resulting in 68.0 and 83.3 on SR and PR, respectively.

In addition, we investigate the overall parameter count and computational complexity of the tracker under different configurations. The naive concatenation operation incurs over 1,000 GFLOPs due to the large number of multimodal input tokens. In contrast, our two core designs significantly reduce computational complexity while simultaneously improving overall performance.

\subsection{Ablation Study}
\label{ablation_study}

\noindent $\bullet$ \textbf{Analysis of Input Data.~}
Table~\ref{tab: Ablation_Studies} reports the tracking performance under four configurations: Event only, RGB only, RGB fused with event frames, and RGB fused with event voxels. Using only event frames yields limited results (SR 58.3, PR 71.1) due to sparsity and noise, whereas RGB frames alone achieve substantially higher performance, underscoring the richness of appearance cues. Fusing RGB frames with event frames improves tracking performance, achieving an SR of 67.5 and a PR of 82.4. Incorporating event voxels with RGB increases accuracy to an SR of 68.0 and a PR of 83.3, demonstrating that their rich motion cues effectively complement RGB frames to improve tracking performance.

\noindent $\bullet$ \textbf{Analysis of Different Fusion Methods.~}
Table~\ref{tab: Ablation_Studies} presents a comparison between our proposed decoupled amplitude–phase attention aggregation method and two commonly used feature fusion methods. In the addition-based experiment, the event voxel is first aligned with the RGB image via a convolutional layer and then directly added to it before being passed through the projection layer. In contrast, the concatenation-based experiment performs token-level concatenation after the projection layer. It is important to note that both baseline experiments retain the MGSS module. The results indicate that our frequency-domain amplitude–phase fusion strategy more effectively incorporates informative cues from the event modality into the RGB modality, ultimately leading to superior tracking performance.

\begin{table}
\center
\small     
\caption{Ablation Studies on the COESOT dataset.} 
\label{tab: Ablation_Studies}
\resizebox{0.95\columnwidth}{!}{ 
\begin{tabular}{l|lll}
\rowcolor{gray!20}
\hline \toprule [0.5 pt] 
\textbf{\# Input Data}    &\textbf{SR}   & \textbf{PR}   \\
\hline
\text{1. Event only}     &58.3     &71.1        \\
\text{2. RGB only}     &66.4   &81.2    \\
\text{3. RGB Frame \& Event Frame}      &67.5   &82.4      \\
\textbf{4. RGB Frame \& Event Voxel}      & \textbf{68.0}     & \textbf{83.3}      \\
\hline 
\rowcolor{gray!20}
\textbf{\# Fusion Methods}    &\textbf{SR}   & \textbf{PR}    \\
\hline
\text{1. Addition}    &67.6   &82.5     \\ 
\text{2. Concatenation}      &67.7    &82.6        \\
\textbf{3. Amplitude and Phase Attention}    & \textbf{68.0}     & \textbf{83.3}      \\
\hline 
\rowcolor{gray!20}
\textbf{\# Different ViT Modules}    &\textbf{SR}   & \textbf{PR}    \\
\hline
\text{1. Standard ViT }     &67.4    &82.7                \\ 
\text{2. Diff-ViT}       &67.6     &83.1        \\ 
\textbf{3. Diff-FFT ViT}  & \textbf{68.0}     & \textbf{83.3}        \\ 
\hline
\rowcolor{gray!20}
\textbf{\# Spatial Sparsification Methods}    &\textbf{SR}   & \textbf{PR}    \\
\hline
\text{1. Random Drop }                   &66.8     &82.0       \\ 
\text{2. Candidate elimination}        &67.4     &82.5        \\ 
\text{3. DynamicViT}        &67.2     &82.4        \\ 
\textbf{4. Motion-Guided}          & \textbf{68.0}     & \textbf{83.3}     \\
\hline 
\rowcolor{gray!20}
\textbf{\# Adaptive Top-$K$ Functions}    &\textbf{SR}   & \textbf{PR}    \\
\hline
\text{1. ${K_{adp}} = K_{max} + (K_{max}-K_{min}) \cdot (-{\beta x})$ }             &67.5     &82.5          \\ 
\text{2. ${K_{adp}} = K_{min} + (K_{max}-K_{min}) \cdot (1 + x)^ {-\beta} $  }              &67.2    &82.0     \\ 
\textbf{3. ${K_{adp}} = K_{min} + (K_{max}-K_{min}) \cdot e^{-\beta x}$ }     & \textbf{68.0}     & \textbf{83.3}       \\ 
\hline \toprule [0.5 pt] 
\end{tabular}
}
\end{table}

\noindent $\bullet$ \textbf{Analysis of Diff-FFT ViT Module.~}
We further analyze the advantages of our Diff-FFT ViT module. As shown in Table~\ref{tab: Ablation_Studies}, the standard ViT achieves an SR of 67.4 and a PR of 82.7, serving as a strong baseline. Diff-ViT further improves these results, reaching an SR of 67.6 and a PR of 83.1. In contrast, our Diff-FFT ViT extends Diff-ViT by performing differential attention in the frequency domain and applying a Gaussian window to suppress noise, thereby yielding the best overall performance. This performance gain suggests that Fourier-domain interactions enable more precise extraction of target-related motion cues while effectively mitigating the influence of noise on critical features.

\noindent $\bullet$ \textbf{Analysis of Spatial Sparsification Methods.~}
This work adopts a motion-guided adaptive spatial sparsification approach to reduce spatial redundancy. In addition, we compare it with other commonly used sparsification methods, as summarized in Table~\ref{tab: Ablation_Studies}. It can be observed that randomly dropping some tokens inevitably results in the loss of critical information, leading to performance degradation. Candidate elimination and dynamic ViT methods can improve the accuracy of redundant token filtering to some extent. Finally, our approach achieves the best performance, demonstrating that motion information from events can more accurately reflect the spatial distribution pattern of the scene, thereby enabling a more robust and effective selection of informative tokens.

\begin{figure}
\includegraphics[width=\linewidth]{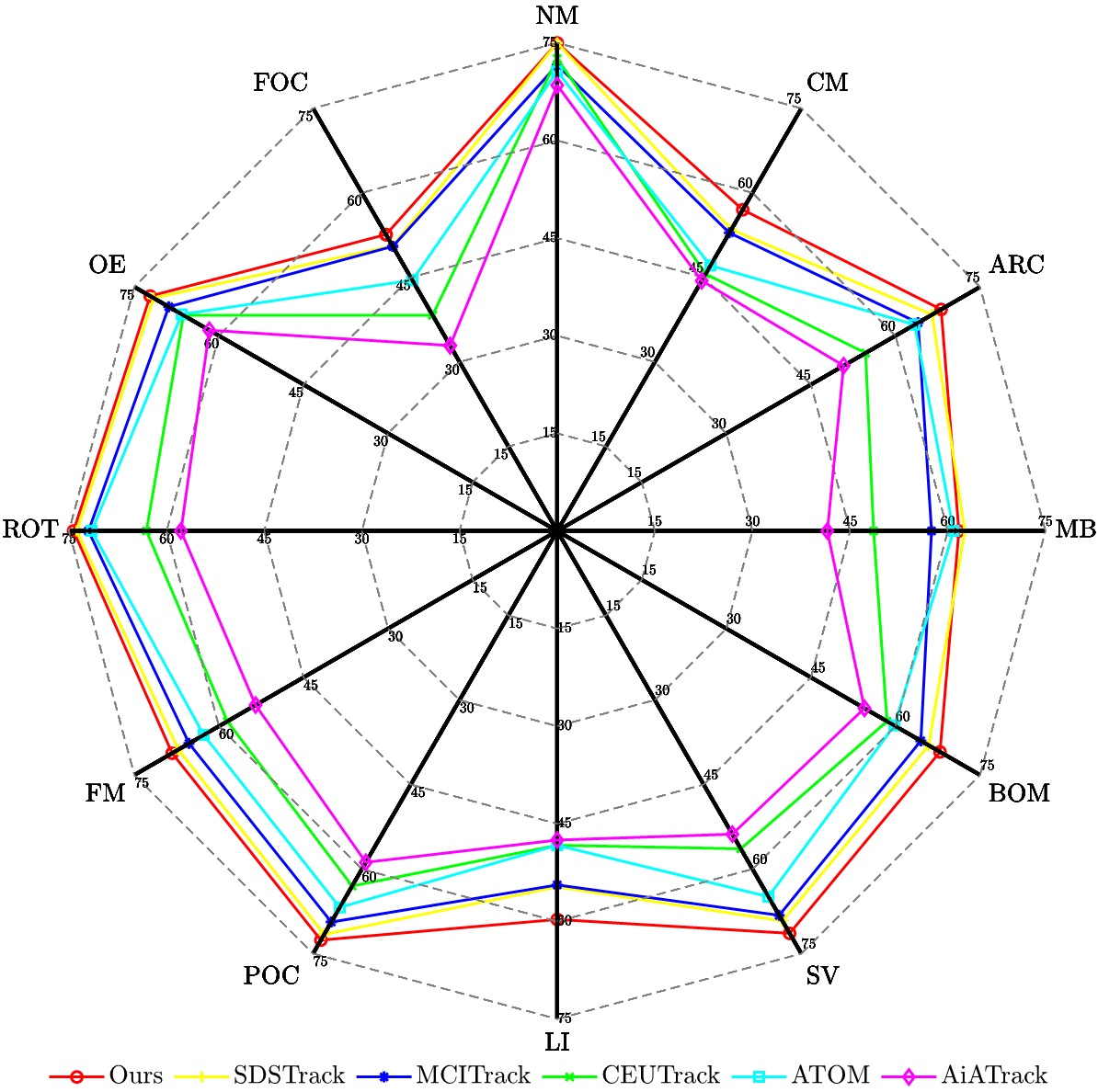}
\caption{Tracking results (SR) under each challenging factor.} 
\label{attributeResults}
\end{figure}

\noindent $\bullet$ \textbf{Analysis of Adaptive Top-$K$ Functions.~}
To dynamically adjust the number of retained tokens, we explore three adaptive Top‑$K$ selection functions based on motion cues. As shown in Table~\ref{tab: Ablation_Studies}, the first variant, which linearly decreases $K_{\mathrm{adp}}$, achieves an SR of 67.5 and PR of 82.5. The second, using a power-law decay $(1 + x)^{-\beta}$, performs slightly worse (SR: 67.2, PR: 82.0), indicating limited adaptation under complex backgrounds. Our third variant, based on an exponential decay $e^{-\beta x}$, achieves the best results, enabling smoother token pruning. These results show that exponential-based adaptive sparsification effectively balances efficiency and tracking accuracy for RGB-Event object tracking.

\begin{figure}
\center
\includegraphics[width=0.48\textwidth]{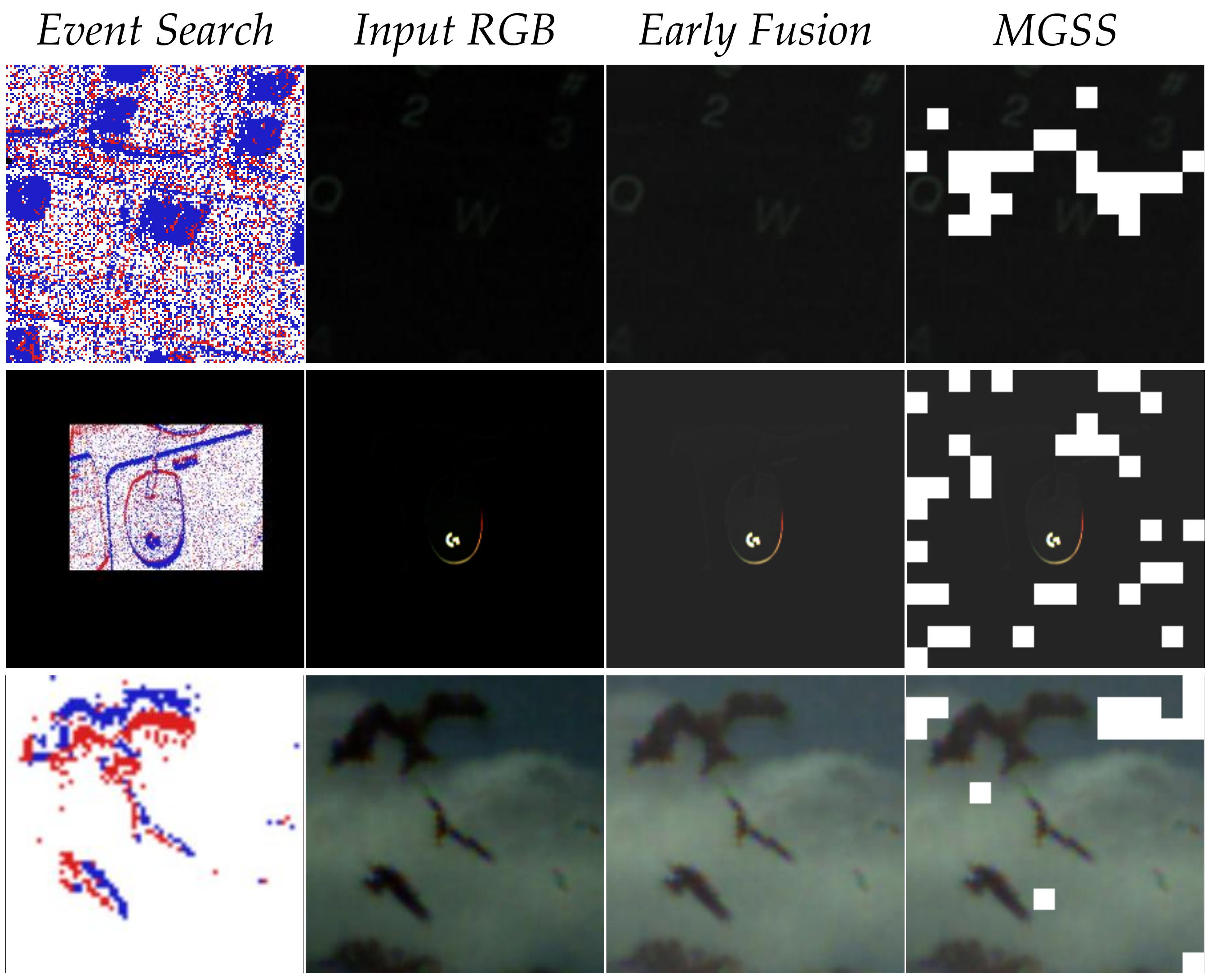}
\caption{Visualization results of our early fusion and motion-guided spatial sparsification strategy.}  
\label{early_fusion&mgss}
\end{figure}

\noindent $\bullet$ \textbf{Success Rate Under Challenging Attributes.~}
We further evaluate our proposed method on the COESOT dataset across 12 challenging attributes, providing a comprehensive assessment of its robustness and adaptability across diverse tracking scenarios.
As shown in Fig.~\ref{attributeResults}, our method achieves the highest success rate on 11 out of 12 attributes compared to SDSTrack, MCITrack, CEUTrack, ATOM, and AiATrack. In particular, it demonstrates a clear advantage in low-illumination (LI) scenarios, indicating that the proposed decoupled amplitude and phase attention effectively integrates high dynamic range information from the event data. Furthermore, its superior performance in background object motion (BOM), full occlusion (FOC), and camera motion (CM) scenarios validates that the motion-guided spatial sparsification efficiently removes redundant noise while focusing on target-relevant cues. These results demonstrate that our model provides highly reliable tracking performance in various real-world scenarios, effectively addressing dynamic environmental changes and complex target behaviors.

\begin{figure}
\center
\includegraphics[width=0.44\textwidth]{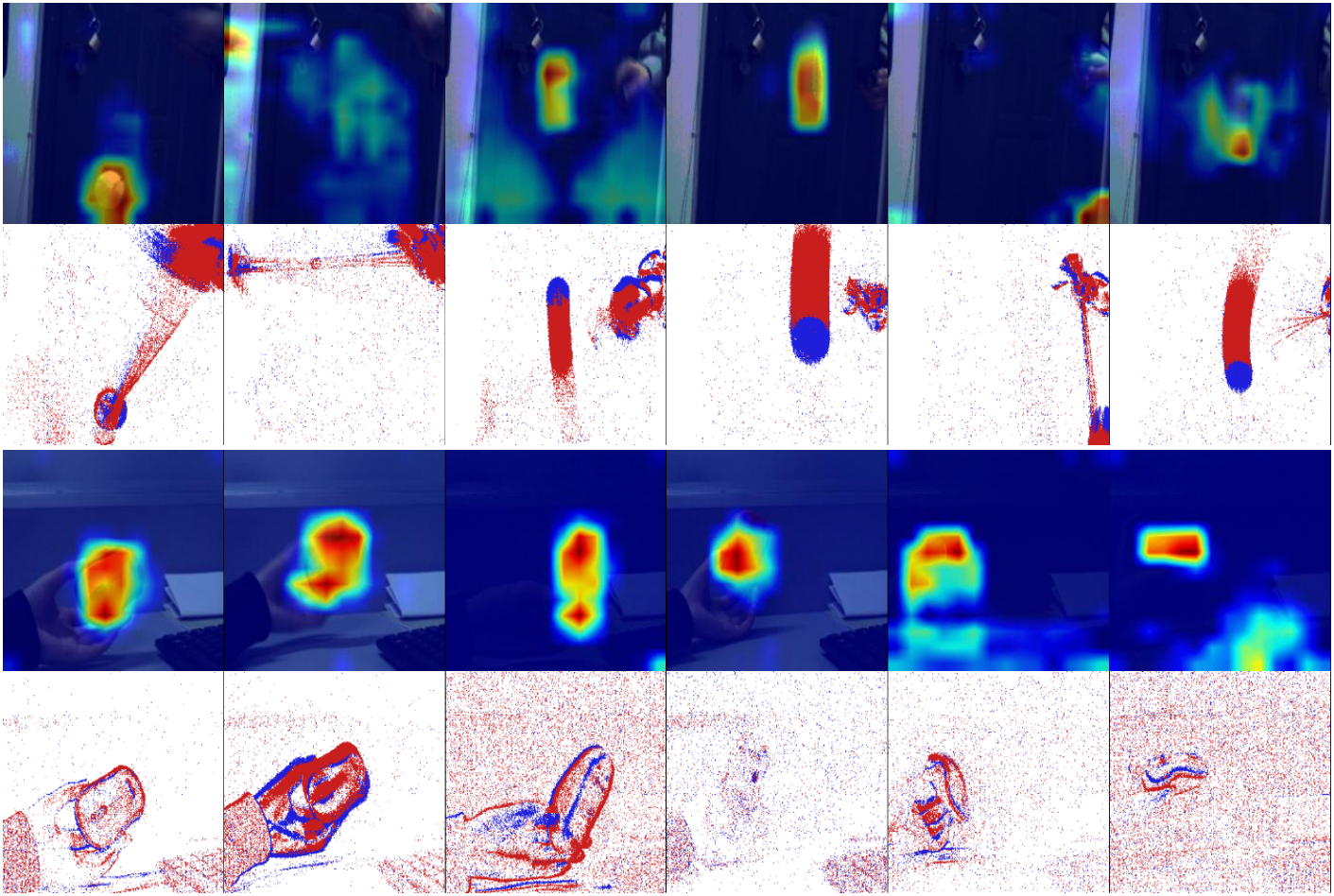}
\caption{Visualization of the attention activation maps generated by our method.}  
\label{attention_map}
\end{figure} 

\begin{figure}
\center
\includegraphics[width=0.44\textwidth]{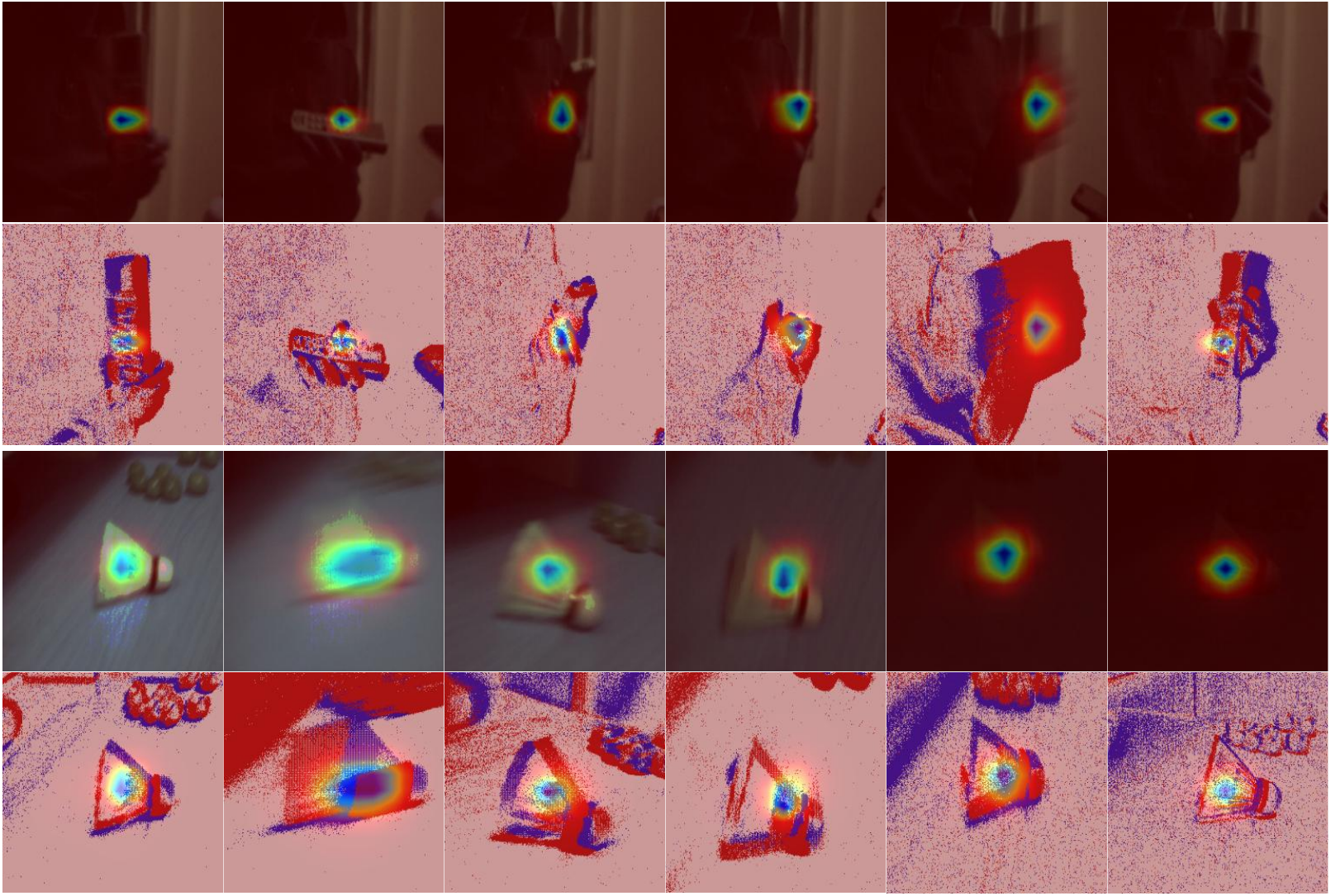}
\caption{Visualization of the response maps generated by our method.}  
\label{response_map}
\end{figure}

\begin{figure*}
\center
\includegraphics[width=0.9\textwidth]{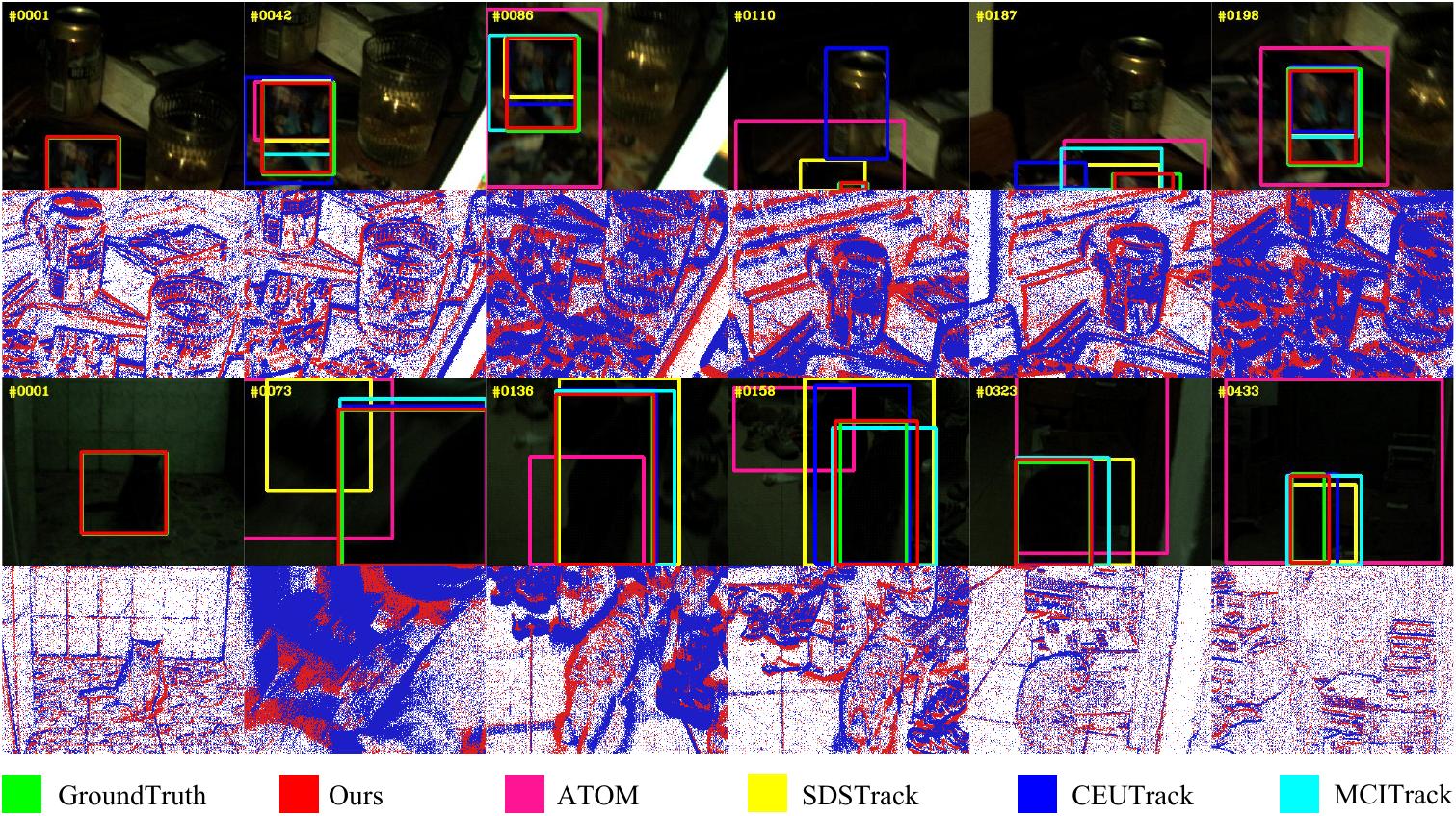}
\caption{Visualization of tracking results produced by our method and other SOTA trackers.}  
\label{trackingResults}
\end{figure*}

\subsection{Efficiency Analysis} 
Table~\ref{CAResults} compares the parameter count and computational complexity of different model components. Compared with conventional multimodal learning methods based on feature concatenation, our approach increases the parameter count by only 2.9M, while significantly reducing computational complexity. This improvement primarily stems from our proposed decoupled amplitude and phase attention aggregation strategy, which halves the number of input tokens fed into the backbone during early fusion. Simultaneously, motion-guided spatial sparsification further reduces spatial redundancy. Together, these mechanisms substantially alleviate the computational burden of the backbone while preserving effective utilization of multimodal information. In addition, we report the inference tracking speed of our framework on the FELT dataset, as shown in Table~\ref{tab:felt_results}. Our model achieves 27 FPS, enabling effective real-time tracking performance.

\subsection{Visualization}

\noindent $\bullet$ \textbf{Results of Early Fusion and Motion-Guided Spatial Sparsification.}
In addition to the quantitative results, in this section, we present several qualitative results to facilitate a clearer understanding of our framework. As illustrated in Fig.~\ref{early_fusion&mgss}, we visualize the enhancement effects of the early fusion method, namely the decoupled amplitude and phase attention module, on the RGB inputs, as well as the results of the motion-guided spatial sparsification (MGSS) module. The visualization results show that the early fusion strategy effectively enhances the original RGB images by incorporating high-frequency information from the event modality, thereby alleviating tracking challenges in complex scenarios. Furthermore, MGSS dynamically suppresses spatially redundant background regions while preserving target-relevant tokens, which improves the model’s capability to discriminate between foreground and background.

\noindent $\bullet$ \textbf{Attention Maps and Response Maps.}
As shown in Fig.~\ref{attention_map}, we visualize the attention activation maps produced by our model. Regions with colors closer to red indicate higher attention weights, while blue regions correspond to areas receiving little attention. It can be seen that our model consistently and accurately focuses on the template target across various challenging scenarios. Furthermore, Fig.~\ref{response_map} presents the final response maps generated by the tracker, where deep blue regions denote high response values. Even in challenging conditions such as target rotation and illumination variations, our method is able to generate strong and precise responses at the target locations within the search region. These results further demonstrate the robustness of our approach.

\noindent $\bullet$ \textbf{Tracking Results.}
In addition, we provide detailed qualitative comparisons of tracking results to facilitate a deeper understanding of our framework. As shown in Fig.~\ref{trackingResults}, we compare the tracking results of our method with several other SOTA trackers, including ATOM, SDSTrack, CEUTrack, and MCITrack, on the COESOT dataset. In video sequences with low illumination and fast motion, our method generates bounding boxes that align most closely with the ground truth. By contrast, the compared trackers often drift to background regions or nearby distractors, and occasionally misidentify the target when the event stream becomes dense. These results further demonstrate that our tracker effectively exploits complementary frame and event cues for more robust RGB-Event-based object tracking.

\subsection{Limitation Analysis} 
Although our framework effectively integrates RGB and event modalities for visual object tracking, it still faces the following two limitations:
(1) The issue of imbalanced multimodal learning still persists, as the model may overly rely on one modality. This can suppress the feature representation of weaker modalities, leading to insufficient learning of multimodal information. A potential solution is to dynamically assign modality-specific weights during training, enabling weaker modalities to receive more attention.
(2) The current framework lacks tailored adaptation strategies for individual challenging scenarios, which may cause certain challenge attributes that degrade overall performance to be overlooked. To address this limitation, a promising future direction is to develop an attribute-based multi-expert strategy, where each expert is specialized in handling specific challenges, thereby further enhancing the algorithm’s robustness.

\section{Conclusion} 
In this work, we propose APMTrack, an effective framework for RGB-Event multimodal visual object tracking. To overcome the limitations of conventional feature-level fusion, we first introduce a decoupled amplitude and phase attention module. In the frequency domain, this module decomposes the amplitude and phase components of the RGB and event modalities. High-frequency information from the event modality is then incorporated into the RGB modality. This not only strengthens the feature representation of the RGB modality but also substantially reduces the computational burden of the backbone network.
In addition, we propose a motion-guided spatial sparsification strategy. By leveraging the motion-sensitive properties of event cameras, this strategy models the relationship between target-relevant motion cues and the spatial probability distribution, adaptively filtering out redundant background regions while enhancing target-relevant features. Consequently, these two modules reduce computational complexity and improve tracking performance, striking a balance between accuracy and efficiency and pushing forward the field of RGB-Event visual object tracking.

\section*{Acknowledgment} 
This work was supported by the National Natural Science Foundation of China (62102205, 62576004, 62572043), Anhui Provincial Natural Science Foundation-Outstanding Youth Project (2408085Y032), Natural Science Foundation of Anhui Province (2408085J037), and Beijing Natural Science Foundation (No. JQ24024). 
The authors acknowledge the High-performance Computing Platform of Anhui University for providing computing resources.

\small{ 
\bibliographystyle{IEEEtran}
\bibliography{reference}
}

\end{document}